# Exact Maximum Margin Structure Learning of Bayesian Networks

Robert Peharz                                                                  robert.peharz@tugraz.at
Franz Pernkopf                                                                       pernkopf@tugraz.at
Signal Processing and Speech Communication Laboratory, Graz University of Technology, Austria

## Abstract

Recently, there has been much interest in finding globally optimal Bayesian network structures. These techniques were developed for generative scores and can not be directly extended to discriminative scores, as desired for classification. In this paper, we propose an exact method for finding network structures maximizing the probabilistic soft margin, a successfully applied discriminative score. Our method is based on branch-and-bound techniques within a linear programming framework and maintains an any-time solution, together with worst-case sub-optimality bounds. We apply a set of order constraints for enforcing the network structure to be acyclic, which allows a compact problem representation and the use of general-purpose optimization techniques. In classification experiments, our methods clearly outperform generatively trained network structures and compete with support vector machines.

## 1. Introduction

Bayesian networks (BNs) are an important type of probabilistic graphical models (Pearl, 1988; Koller & Friedmann, 2009) and specify a probability distribution over a set of random variables (RVs). They make use of a directed acyclic graph (DAG), with nodes corresponding to the RVs, representing the factorization of the joint distribution. Learning the structure of Bayesian networks from data can be cast as optimization problem, where the goal is to find a DAG maximizing some score function. This is a combinatorial problem and known to be NP-hard in general (Chickering, 1996). Therefore, most approaches to learn the BN structure are approximative or greedy heuristics. Recently, there has been much interest in *exact* structure learning, i.e. in finding globally optimal DAGs. Koivisto et al. (2004) use a dynamic programming approach for efficiently summing over all variable orders, leading to exponential (rather than super-exponential) run-time. Further development of this approach can be found in (Silander & Myllymäki, 2006; Parviainen & Koivisto, 2009). Due to exponential run-time, these methods are currently restricted to approximately 30-50 variables. Alternatively to dynamic programming, branch-and-bound (B&B) techniques have been exploited for exact structure learning (de Campos et al., 2009; de Campos & Ji, 2011; Jaakkola et al., 2010). In comparison to dynamic programming, these techniques offer the advantage of an *any-time* solution, i.e. as soon as some feasible solution has been found, the algorithm can be interrupted and returns the currently best solution, together with a worst-case sub-optimality bound. However, if the algorithm is kept running, it eventually finds a globally optimal solution.

These methods have been developed for *generative* structure learning, i.e. they aim to maximize generative scores such as MDL/BIC (Rissanen, 1978; Suzuki, 1993) or BDe (Buntine, 1991; Cooper & Herskovits, 1992; Heckerman et al., 1995). These scores are decomposable, i.e. they can be written as a sum over local scores, one for each random variable. On the other hand, when the learned BN shall be used as classifier, we aim to maximize a *discriminative* score, such as (parameter penalized) conditional likelihood, classification rate, or a recently proposed probabilistic margin formulation (Guo et al., 2005; Pernkopf et al., 2012). Using a discriminative criterion typically leads to classifiers with higher accuracy, especially, when the selected model class does not capture the underlying data distribution. However, since these discriminative BN scores are not decomposable, the discussed methods for exact generative structure learning cannot be directly applied.

In this paper, we propose an exact method for learning a BN structure maximizing the probabilistic margin. For this purpose, we use concepts developed in (de Campos et al., 2009; Jaakkola et al., 2010; Cussens, 2011), leading to a formulation as *mixed integer linear program* (MILP). For solving a MILP, the problem is relaxed in a linear program (LP), and a B&B method is used to enforce integrality. Therefore, our work falls within the line of research using B&B for exact structure learning, but maximizing a discriminative criterion. Similar as in (Cussens, 2010), we use a set of order constraints to enforce acyclicity in the directed graph, rather than the cluster constraints used in (Jaakkola et al., 2010). The advantage of this formulation is, that for $N$ RVs, we only require $N^2 - N$ linear constraints for enforcing acyclicity, rather than super-exponentially many cluster constraints. Consequently, we are able to compactly represent our problem and to use powerful general-purpose solvers for structure learning. Although we learn a discriminative BN structure in order to obtain good classifiers, we still use maximum likelihood parameters. Therefore, the resulting BN consistently approximates the true underlying distribution and is suitable for all kinds of inference scenarios.

Similarly as in (Guo et al., 2005), the margin formulation needs one linear constraint per training sample and per competing class. This can render the approach infeasible for problems with many training samples and many class values. Therefore, as a second contribution, we propose a binary margin formulation, which can be interpreted as a local (sample-wise) one-versus-all classification scheme. The problem size using the binary margin does not depend on the number of classes, which is computationally beneficial for problems with many classes. For binary classification problems, the two margin formulations are equivalent. For multi-class problems, we empirically shown that the binary margin classifier competes with the original max-margin structure. We perform classification experiments on 31 datasets, and compare our algorithms with naive Bayes, tree-augmented naive Bayes, generative learned BNs and support vector machines (SVMs). The max-margin structures outperform the other BN classifiers on most datasets, and compete with support vector machines.

The paper is organized as follows. In section 2, we review BNs and introduce our notation, and in section 3, we review related work. We present our method for max-margin structure learning in section 4. The binary margin formulation is introduced in 5. In section 6 we present our experiments and section 7 concludes the paper.

## 2. Background and Notation

Throughout the paper we assume discrete RVs, where plain capital letters denote single RVs and capital boldface letters represent sets of RVs. The set of states which can be assumed by RV $X$ is denoted as $\mathbf{val}(X)$, and we define $\mathbf{sp}(X) = |\mathbf{val}(X)|$. For simplicity of notation, we identify the states of an RV $X$ with natural numbers, i.e. $\mathbf{val}(X) = \{1, \ldots, \mathbf{sp}(X)\}$. However, we do not assume a particular ordering or interpretation of these states. Furthermore, we use $\mathbf{val}(\mathbf{X})$ to denote the set of possible joint states of a set of RVs $\mathbf{X}$, and let $\mathbf{sp}(\mathbf{X}) = |\mathbf{val}(\mathbf{X})|$. Lower-case plain letters represent values or states of RVs, e.g. $x$ is a value of RV $X$. Similarly, lower-case boldface letter represent joint states of variable sets, e.g. $\mathbf{x}$ is a state of RV set $\mathbf{X}$. When $\mathbf{y}$ is a state of $\mathbf{Y}$, and $\mathbf{X}$ is a subset of $\mathbf{Y}$, then $\mathbf{y}_{(\mathbf{X})}$ denotes the corresponding state of $\mathbf{X}$.

A BN $\mathcal{B}$ over a set of $N$ RVs $\mathbf{X} = \{X_1, \ldots, X_N\}$ is defined as a tuple $\langle \mathcal{G}, \Theta \rangle$, where $\mathcal{G}$ is a DAG, with nodes corresponding to the RVs in $\mathbf{X}$. The set of parents of $X_i$ according to $\mathcal{G}$ is denoted as $\mathbf{Pa}_i$. The set $\Theta = \{\theta^1, \ldots, \theta^N\}$ contains parameter sets $\theta^i = \{\theta^i_{j|\mathbf{h}}, \forall j \in \mathbf{val}(X_i), \forall \mathbf{h} \in \mathbf{val}(\mathbf{Pa}_i)\}$ for each variable $X_i$, parameterizing a conditional probability distribution: $P(X_i = j | \mathbf{Pa}_i = \mathbf{h}; \theta^i) = \theta^i_{j|\mathbf{h}}$. A BN defines a probability distribution over the RVs, according to

$$P_{\mathcal{B}}(\mathbf{X} = \mathbf{x}) = \prod_{i=1}^{N} \prod_{j=1}^{\mathbf{sp}(X_i)} \prod_{\mathbf{h} \in \mathbf{val}(\mathbf{Pa}_i)} {\theta^i_{j|\mathbf{h}}}^{\nu^i_{j|\mathbf{h}}}, \quad (1)$$

where $\nu^i_{j|\mathbf{h}}$ is the indicator function $\mathbb{1}(x_i = j \text{ and } \mathbf{x}_{(\mathbf{Pa}_i)} = \mathbf{h})$. For classification, one of the variables in $\mathbf{X}$ represents the class variable $C$, where without loss of generality, we assume that $X_1 = C$ and let $\mathbf{Z} = \{X_2, \ldots, X_N\}$. Let $\mathcal{D} = \{\mathbf{x}^1, \ldots \mathbf{x}^M\}$ be a collection of $M$ i.i.d. samples drawn from some unknown distribution. It is well known that for a fixed BN structure $\mathcal{G}$, the maximum likelihood (ML) parameters $\hat{\Theta}$ are given as

$$\hat{\theta}^i_{j|\mathbf{h}} = \frac{n^i_{j|\mathbf{h}}}{n^i_{\mathbf{h}}}, \quad (2)$$

where $n^i_{j|\mathbf{h}} = \sum_{m=1}^{M} \nu^{i,m}_{j|\mathbf{h}}$, $n^i_{\mathbf{h}} = \sum_{j=1}^{\mathbf{sp}(X_i)} n^i_{j|\mathbf{h}}$, and $\nu^{i,m}_{j|\mathbf{h}} = \mathbb{1}(x^m_i = j \text{ and } \mathbf{x}^m_{(\mathbf{Pa}_i)} = \mathbf{h})$. We can regularize these parameters using Laplace-smoothing, by replacing $n^i_{j|\mathbf{h}}$ with $n^i_{j|\mathbf{h}} + 1$. The smoothed parameters are also consistent ML estimators, although biased towards a uniform distribution.

## 3. Related Work

We adopt the framework developed in (Jaakkola et al., 2010), where the aim was to maximize a generative score such as MDL/BIC or BDe. For each variable $X_i$, we identify a set of possible parent-sets $\mathcal{S}_i = \{\mathbf{S}_{i,1}, \ldots \mathbf{S}_{i,Q_i}\}$, where $Q_i = |\mathcal{S}_i|$ and each $\mathbf{S}_{i,j} \subseteq \mathbf{X} \setminus X_i$, $j \in \{1, \ldots, Q_i\}$. A specific network structure $\mathcal{G}$ is represented by selecting a single parent-set from each $\mathcal{S}_i$, $i \in \{1, \ldots, N\}$. The sets $\mathcal{S}_i$ have to be reasonable large to represent a variety of solutions, while being reasonable small, such that the algorithm remains tractable. In (de Campos et al., 2009), a pruning strategy was presented, for a-priori excluding all parent-sets, which can not occur in an optimal DAG. Since not every combination of parent-sets yields an *acyclic* graph, additional acyclicity constraints have to be imposed. More formally, let $\boldsymbol{\omega}_i = (\omega_{i,1}, \ldots, \omega_{i,Q_i})^T$ be a vector of pre-computed local scores, where $\omega_{i,j}$ is the local score for $\mathbf{S}_{i,j}$. Furthermore, let $\boldsymbol{\eta}_i = (\eta_{i,1}, \ldots, \eta_{i,Q_i})^T$ be the parent-set indicator vector, which contains exactly one 1, indicating the selected parent-set in $\mathcal{S}_i$, and which is 0 elsewhere. All $\boldsymbol{\eta}_i$ are stacked into a single vector $\boldsymbol{\eta} = (\boldsymbol{\eta}_1^T, \ldots, \boldsymbol{\eta}_N^T)^T$, and similarly $\boldsymbol{\omega} = (\boldsymbol{\omega}_1^T, \ldots, \boldsymbol{\omega}_N^T)^T$. Let $\mathcal{P}$ be the convex hull of all vectors $\boldsymbol{\eta}$ which represent valid DAGs. Consequently, all vertices of $\mathcal{P}$ represent DAGs, and as easily shown, all vectors $\boldsymbol{\eta}$ representing cyclic graphs are not elements of $\mathcal{P}$. Generative structure learning is cast as the LP

$$\begin{aligned} \text{maximize} \quad & \boldsymbol{\omega}^T \boldsymbol{\eta} \\ \text{s.t.} \quad & \boldsymbol{\eta} \in \mathcal{P}. \end{aligned} \quad (3)$$

Since there is always an optimal solution in some vertex of $\mathcal{P}$, and since each vertex represents a DAG, we are in principle able to recover an optimal structure by solving (3). Note that $\mathcal{P}$ has super-exponentially many facets in the number of variables, which, in agreement with theory (Chickering, 1996), makes the problem hard. Unfortunately, no representation of $\mathcal{P}$ via linear inequalities is known. Therefore, in (Jaakkola et al., 2010; Cussens, 2011) the constraint in problem (3) was replaced with

$$\eta_{i,j} \geq 0 \qquad i \in \{1, \ldots, N\}, j \in \{1, \ldots, Q_i\} \quad (4)$$

$$\sum_{j=1}^{Q_i} \eta_{i,j} = 1 \qquad i \in \{1, \ldots, N\} \quad (5)$$

$$\eta_{i,j} \in \mathbb{Z} \qquad i \in \{1, \ldots, N\}, j \in \{1, \ldots, Q_i\} \quad (6)$$

$$\sum_{i: X_i \in \mathbf{C}} \sum_{j=1}^{Q_i} \eta_{i,j} \mathbb{1}(\mathbf{S}_{i,j} \cap \mathbf{C} = \emptyset) \geq 1 \qquad \forall \mathbf{C} \subseteq \mathbf{X} \quad (7)$$

The constraints (4), (5) and (7) were used as approximation for $\mathcal{P}$. However, since the solution of the LP might be fractional, the integrality constraint (6) is required. The constraints (4)-(6) can be interpreted as the constraint "$\boldsymbol{\eta}$ represents a directed graph", since they enforce that exactly one entry in each $\boldsymbol{\eta}_i$ is 1, and all others are 0. Constraints (7) enforce acyclicity, since they enforce that for each cluster $\mathbf{C} \subseteq \mathbf{X}$, there is at least one variable $X_i \in \mathbf{C}$ whose parent-set is either outside $\mathbf{C}$, or which is empty. Thus, constraints (4)-(7) force $\boldsymbol{\eta}$ to represent a DAG. Note that the problem has super-exponentially many constraints. Jaakkola et al. (2010) solve the relaxed problem in the dual, where each in-active cluster constraint corresponds to a zero dual variable, and Cussens (2011) uses a cutting plane approach, iteratively adding violated cluster constraints.

As already noted, it is the decomposability of generative scores which yields the linear objective $\boldsymbol{\omega}^T \boldsymbol{\eta}$ in (3), leading to the LP formulation. Since discriminative scores are usually not decomposable, the LP approach cannot be directly applied. However, in the next section we derive an exact MILP formulation for the so-called probabilistic soft margin. The margin $\delta^m$ of the $m^{\text{th}}$ sample is defined as (Guo et al., 2005):

$$\delta^m = \frac{P_{\mathcal{B}}(c^m | \mathbf{z}^m)}{\max_{c \neq c^m} P_{\mathcal{B}}(c | \mathbf{z}^m)} = \frac{P_{\mathcal{B}}(c^m, \mathbf{z}^m)}{\max_{c \neq c^m} P_{\mathcal{B}}(c, \mathbf{z}^m)}. \quad (8)$$

When $\delta^m > 1$, then the $m^{\text{th}}$ sample is correctly classified, and when $\delta^m < 1$, it is wrongly classified. Motivated by SVMs, Pernkopf et al. (2012) defined a soft margin (SM) using the hinge loss:

$$SM(\mathcal{B}) = \sum_{m=1}^{M} \min(\log \delta^m, \gamma). \quad (9)$$

The log-margin of each sample contributes linearly to the overall score. To avoid that the score is mainly determined by a few samples with overly large margin, the sample margins are limited with the hinge function $\min(\cdot, \gamma)$. The parameter $\gamma$, which is obtained by cross validation, has the interpretation as "desired log-margin" for each sample. In (Pernkopf et al., 2012) this score was used for parameter learning using a conjugate gradient method, and in (Pernkopf et al., 2011; Pernkopf & Wohlmayr, 2012) the same score was used for inexact BN structure learning, based on greedy hill-climbing and simulated annealing.

## 4. Max-Margin Structure Learning

We aim to find a BN structure $\mathcal{G}$ *globally* maximizing the SM score in (9). First, we restrict the maximal number of parents for each variable, to obtain a tractable number of parent-sets. For a variable $X_i \neq C$, we only need to consider the empty parent set, and parent-sets containing $C$, since all other parent-sets do not influence the margin.[1] We further assume Laplace-smoothed ML parameters while learning the structure. Firstly because simultaneous learning of max-margin structure and parameters would render our approach intractable. Secondly, by using generative parameters, the resulting BN can still be interpreted as generative model, although its structure is determined discriminatively.

Using the notation introduced in section 3, we can expand the BN distribution (1) according to

$$P_\mathcal{B}(\mathbf{x}) = \prod_{i=1}^{N} \prod_{j=1}^{\mathbf{sp}(X_i)} \prod_{k=1}^{Q_i} \prod_{\mathbf{h} \in \mathbf{val}(\mathbf{S}_{i,k})} \theta_{j|\mathbf{h}}^{i,k \; \nu_{j|\mathbf{h}}^{i,m,k} \eta_{i,k}}, \quad (10)$$

where $\theta_{j|\mathbf{h}}^{i,k}$ are the ML parameters when $\mathbf{S}_{i,k}$ is the parent-set of variable $X_i$, and $\nu_{j|\mathbf{h}}^{i,m,k} = \mathbb{1}(x_i^m = j \text{ and } \mathbf{x}_{(\mathbf{S}_{i,k})}^m = \mathbf{h})$. Clearly, (10) represents the same distribution as (1), where the structure $\mathcal{G}$ is explicitly encoded with $\boldsymbol{\eta}$. Inserting (10) in the margin definition (8) and taking the log, gives

$$\log \delta^m = \sum_{i=1}^{N} \sum_{j=1}^{\mathbf{sp}(X_i)} \sum_{k=1}^{Q_i} \sum_{\mathbf{h}} \nu_{j|\mathbf{h}}^{i,m,k} \eta_{i,k} \log \theta_{j|\mathbf{h}}^{i,k} -$$

$$\max_{c \neq c^m} \left[ \sum_{i=1}^{N} \sum_{j=1}^{\mathbf{sp}(X_i)} \sum_{k=1}^{Q_i} \sum_{\mathbf{h}} \nu_{j|\mathbf{h}}^{i,m,k,c} \eta_{i,k} \log \theta_{j|\mathbf{h}}^{i,k} \right]$$

$$= \min_{c \neq c^m} \left[ \sum_{i=1}^{N} \sum_{k=1}^{Q_i} \eta_{i,k} \, \alpha(i,k,m,c) \right], \quad (11)$$

where $\nu_{j|\mathbf{h}}^{i,m,k,c}$ is defined as $\nu_{j|\mathbf{h}}^{i,m,k}$, but where the class value in $\mathbf{x}^m$ is replaced with the value $c$. The coefficient $\alpha(i,k,m,c)$ is given as

$$\alpha(i,k,m,c) = \sum_{j=1}^{\mathbf{sp}(X_i)} \sum_{\mathbf{h}} \nu_{j|\mathbf{h}}^{i,m,k} \log \theta_{j|\mathbf{h}}^{i,k} - \nu_{j|\mathbf{h}}^{i,m,k,c} \log \theta_{j|\mathbf{h}}^{i,k}. \quad (12)$$

By defining a vector $\boldsymbol{\alpha}_{m,c}$ containing the coefficients $\alpha(i,k,m,c)$ corresponding with the entries of $\boldsymbol{\eta}$, the log-margin in (11) can be written as a minimum over inner products:

$$\log \delta^m = \min_{c \neq c^m} \boldsymbol{\alpha}_{m,c}^T \boldsymbol{\eta}. \quad (13)$$

Using a standard technique from linear programming, we can express the SM score in (9) as follows. We introduce a variable $\tau^m$ for each sample, together with the constraints

$$\tau^m \leq \boldsymbol{\alpha}_{m,c}^T \boldsymbol{\eta}, \quad \forall m, \forall c \neq c^m, \quad (14)$$
$$\tau^m \leq \gamma, \quad (15)$$

and maximize $\sum_{m=1}^{M} \tau^m$. In an optimal LP solution we have $\tau^m = \min(\log \delta^m, \gamma)$, and $\sum_{m=1}^{M} \tau^m$ is precisely the SM score. As in generative structure learning, the DAG constraint could in principle be addressed by constraints (4)-(7). However, in this paper we use a more convenient way to express acyclicity, allowing a compact MILP representation of our problem. Therefore, we replace the cluster constraints (7) with alternative *order constraints*, enforcing a topological ordering among the nodes and thus acyclicity of the resulting graph. We introduce a real-valued *order variable* $o_i$ for each variable $X_i$, which is constrained to $0 \leq o_i \leq \Delta$, where $\Delta$ is some arbitrary positive number. The order constraints are:

$$(1 - a_{i,j})\, 2\Delta + o_j - o_i \geq \frac{\Delta}{N} \quad (16)$$
$$\forall i,j \in \{1,\ldots,N\}, i \neq j,$$

where $a_{i,j} = \sum_{k=1}^{Q_i} \eta_{j,k} \mathbb{1}(X_i \in \mathbf{S}_{j,k})$. The following proposition shows that these constraints enforce acyclicity.

**Proposition 1.** *A vector $\boldsymbol{\eta}$ represents a DAG if and only if there exist some $o_i$, $i \in \{1,\ldots,N\}$, with $0 \leq o_i \leq \Delta$, for some arbitrary $\Delta > 0$, such that constraints (16) and (4)-(6) are fulfilled.*

*Proof.* First we show that when the conditions in the proposition hold, then $\boldsymbol{\eta}$ is necessarily a DAG. Constraints (4)-(6) enforce that $\boldsymbol{\eta}$ represents some directed graph $\mathcal{G}$, since in this case each $\boldsymbol{\eta}_i$ contains exactly one 1 and is 0 elsewhere. It follows that also $a_{i,j}$ is either 1 or 0, and equals an entry of the adjacency matrix of $\mathcal{G}$, indicating an edge from $X_i$ to $X_j$. We switch cases: When $a_{i,j} = 0$, i.e. when there is no edge $X_i \rightarrow X_j$, then (16) yields

$$2\Delta + o_j - o_i \geq 2\Delta + 0 - \Delta = \Delta \geq \frac{\Delta}{N}, \quad (17)$$

and the constraint is fulfilled regardless of $o_i, o_j$. When $a_{i,j} = 1$, i.e. when there is an edge $X_i \rightarrow X_j$, then we have

$$o_j - o_i \geq \frac{\Delta}{N}, \quad (18)$$

which implies that $o_j$ is *strictly larger* than $o_i$. By sorting all $o_i$, we obtain an ordering among the variables (among several $o_i$ with the same value, we pick an arbitrary ordering). Since there can not be an edge $X_i \to X_j$ when $X_i$ comes after $X_j$, this ordering is a topological ordering, and thus the resulting directed graph is acyclic.

---
[1] More precisely, for any $X_i \neq C$, it is equivalent to select the empty parent-set, or to select a parent-set not containing the class variable.

It remains to show that when $\boldsymbol{\eta}$ represents a DAG, then the conditions in proposition 1 hold. When $\boldsymbol{\eta}$ represents a DAG, then it fulfills constraints (4)-(6), since it is a directed graph. Furthermore, we can obtain some topological ordering from the DAG. Let $\bar{o}_i$ be the index of $X_i$ in this ordering. Setting $o_i = (\bar{o}_i - 1)\frac{\Delta}{N}$ fulfills constraints (16) and $0 \leq o_i \leq \Delta$. □

In contrast to the super-exponentially many cluster constraints in (7), we only need $N^2 - N$ linear order constraints and $N$ additional real-valued order variables to enforce acyclicity. Thus, the resulting problem has a more compact representation and can be solved by general-purpose MILP solvers. Similar constraints, using the same mechanism as depicted here, have been proposed for maximum-likelihood pedigree learning (Cussens, 2010). To summarize, our MILP formulation for finding a DAG maximizing the SM in (9), is given as:

$$\begin{array}{ll}
\max. & \sum_{m=1}^{M} \tau^m \\
\text{s.t.} & \tau^m \leq \boldsymbol{\alpha}_{m,c}^T \boldsymbol{\eta}, & \forall m, \forall c \neq c^m \\
& \tau^m \leq \gamma & \forall m \\
& \eta_{i,j} \geq 0 & \forall i, j \\
& \sum_{j=1}^{Q_i} \eta_{i,j} = 1 & \forall i \\
& \eta_{i,j} \in \mathbb{Z} & \forall i, j \\
& (1 - a_{i,j})\, 2\Delta + o_j - o_i \geq \frac{\Delta}{N} & \forall i, j, i \neq j
\end{array} \quad (19)$$

Note that this formulation can immediately be applied to decomposable (i.e. practically all generative) scores. We simply remove the first two constraints (the margin constraints), and replace the objective with the objective of problem (3). Before we present experimental results in section 6, we introduce an alternative margin formulation, leading to a simpler problem.

## 5. Binary Margin Formulation

The main limitation of problem (19) is that we have a constraint $\tau^m \leq \boldsymbol{\alpha}_{m,c}^T \boldsymbol{\eta}$ for each sample and each competing class, i.e. in total $M(\mathbf{sp}(C)-1)$ linear constraints for specifying the margin.[2] One approach would be to reduce the number of samples $M$, by using a representative sub-set of $\mathcal{D}$. Similar techniques have been proposed for clustering and mixture model training, and we plan to address this point in future work. Here, we propose a formulation which only needs $M$ constraints rather than $M(\mathbf{sp}(C)-1)$, which alleviates especially problems with many class values. The basic idea is to employ a one-versus-all classification scheme. When we desire to train a max-margin BN for classifying class $c$ versus all other classes, we replace the parameters used in (10)-(12) with

$$\theta_{j|\mathbf{h}}^{i,k}(c) = \frac{n_{j|\mathbf{h}}^{i,k}(c)}{n_{\mathbf{h}}^{i,k}(c)}, \quad (20)$$

where

$$n_{j|\mathbf{h}}^{i,k}(c) = \sum_{m=1}^{M} \nu_{j|\mathbf{h}}^{i,m,k}(c), \quad n_{\mathbf{h}}^{i,k}(c) = \sum_{j=1}^{\mathbf{sp}(X_i)} n_{j|\mathbf{h}}^{i,k}(c)$$

$$\nu_{j|\mathbf{h}}^{i,m,k}(c) = \begin{cases} \nu_{j|\mathbf{h}}^{i,m,k,1} & \text{if } c^m = c \\ \nu_{j|\mathbf{h}}^{i,m,k,2} & \text{otherwise.} \end{cases}$$

---
[2]The constraint $\tau^m \leq \gamma$ is a simple variable bound and can be treated rather easily.

In words, $\theta_{j|\mathbf{h}}^{i,k}(c)$ are those ML parameters when we re-interpret class value $c$ as class 1 and all other class values as class 2. We could now train a one-versus-all classifier for each $c \in \{1,\ldots,\mathbf{sp}(C)\}$ and combine their decisions for the multi-class problem. However, it is generally unclear how to correctly combine the decisions of one-versus-all classifiers. Furthermore, we now would have to train $\mathbf{sp}(C)$ different classifiers instead of a single one, although each problem would have only $M$ instead of $(\mathbf{sp}(C)-1)M$ margin constraints. Instead, we use the following local, i.e. sample-wise, one-versus-all scheme. Let $\Theta(c)$ be the collection of the parameters according to (20), for all $c \in \{1,\ldots,\mathbf{sp}(C)\}$. Let $P_{\mathcal{B},\Theta(c)}(\mathbf{X})$ be the BN distribution with parameters $\Theta(c)$ and $P_{\mathcal{B},\Theta}(\mathbf{X})$ the BN distribution with original parameters $\Theta$. We define the *binary margin* $\bar{\delta}^m$ of the $m^{\text{th}}$ sample as

$$\bar{\delta}^m = \frac{P_{\mathcal{B},\Theta(c^m)}(1, \mathbf{z}^m)}{P_{\mathcal{B},\Theta(c^m)}(2, \mathbf{z}^m)}. \quad (21)$$

Clearly, $P_{\mathcal{B},\Theta}(c^m, \mathbf{z}^m) = P_{\mathcal{B},\Theta(c^m)}(1, \mathbf{z}^m)$, since we used exactly the same statistics for class $c$ in the original parameterization $\Theta$, and for class 1 in the alternative parameterization $\Theta(c^m)$. When $\mathbf{sp}(C) = 2$, then $\Theta(1)$ and $\Theta(2)$ are simply redundant versions of $\Theta$. In this case we have $P_{\mathcal{B},\Theta}(C \neq c^m, \mathbf{z}^m) = P_{\mathcal{B},\Theta(c^m)}(2, \mathbf{z}^m)$, and $\bar{\delta}^m = \delta^m$, i.e. the margin and the binary margin are *equivalent* for binary classification problems. For $\mathbf{sp}(C) > 2$, $P_{\mathcal{B},\Theta(c^m)}(2, \mathbf{z}^m)$ is an *approximation* for $P_{\mathcal{B},\Theta}(C \neq c^m, \mathbf{z}^m) = \sum_{c' \neq c^m} P_{\mathcal{B},\Theta}(c', \mathbf{z}^m)$, assuming that $P_{\mathcal{B},\Theta(c^m)}(\mathbf{z}^m) \approx P_{\mathcal{B},\Theta}(\mathbf{z}^m)$.

Note that the additional parameters can be stored efficiently, since the parameters for class 1 (for each $\Theta(c)$) are already stored in the original parameter set $\Theta$. Therefore, we need only twice as much memory for storing $\Theta$ and $\Theta(c)$, $c \in \{1, \ldots, \mathbf{sp}(C)\}$, than for storing $\Theta$ alone. Following a similar derivation as for (13), we obtain

$$\log \bar{\delta}^m = \bar{\boldsymbol{\alpha}}_m^T \boldsymbol{\eta}, \tag{22}$$

where $\bar{\boldsymbol{\alpha}}_m$ contains the coefficients (cf. (12))

$$\bar{a}(i, k, m) = \sum_{j=1}^{\mathbf{sp}(X_i)} \sum_{\mathbf{h}} \nu_{j|\mathbf{h}}^{i,m,k,1} \log \theta_{j|\mathbf{h}}^{i,k}(c^m) - \nu_{j|\mathbf{h}}^{i,m,k,2} \log \theta_{j|\mathbf{h}}^{i,k}(c^m). \tag{23}$$

Similar as in (9), we define a soft binary margin $SBM(\mathcal{B}) = \sum_{m=1}^{M} \min(\log \bar{\delta}^m, \gamma)$. The MILP for finding a DAG maximizing the SBM is defined as in (19), except that the constraints

$$\tau^m \leq \boldsymbol{\alpha}_{m,c}^T \boldsymbol{\eta}, \quad \forall m, \forall c \neq c^m \tag{24}$$

are replaced with

$$\tau^m \leq \bar{\boldsymbol{\alpha}}_m^T \boldsymbol{\eta}, \quad \forall m. \tag{25}$$

The alternative parameters $\Theta(c)$ are only needed to obtain the coefficients $\bar{a}(i, k, m)$. For the final BN classifier, we use the original parameters $\Theta$.

## 6. Experiments

We performed classification experiments on 31 dataset obtained from the *UCI machine learning repository* (Frank & Asuncion, 2010). We used the 25 datasets already used in (Friedman et al., 1997), plus six additional datasets: "abalone", "adult", "car", "mushroom", "nursery", and "spambase". These datasets have between 4 and 57 input features and contain between 80 and 45222 samples. For a more detailed information we refer the reader to (Frank & Asuncion, 2010). To estimate the accuracy of the classifiers, a test set was used for the datasets "chess", "letter", "mofn-3-7-10", "satimage", "segment", "shuttle-small", "waveform-21", and the six additional datasets. For the remaining datasets, 5-fold cross-validation was used to estimate the accuracy. Samples with missing features were removed beforehand, and continuous features were discretized using the method described in (Fayyad & Irani, 1993). We compared our methods with naive Bayes (NB), the tree-augmented naive Bayes (TAN) (Friedman et al., 1997) and with a BN with generatively trained structure, using MDL as score function (Suzuki, 1993). Furthermore, we compared with SVMs using a Gaussian kernel, using the LIB-SVM implementation (Chang & Lin, 2011). For the SVM parameters $\sigma$ (width of Gaussian kernel) and $C$ (trade-off factor), we validated all combinations of $\sigma \in \{2^{-5}, 2^{-4}, \ldots, 2^5\}$ and $C \in \{2^{-3}, 2^{-2}, \ldots, 2^5\}$. For our methods, BN structure learning using the SM and SBM, we validated the parameter $\gamma$ (desired log margin), where we used $\gamma = \log\left(\frac{p}{1-p}\right)$, with $p \in \{0.501, 0.6, 0.7, 0.8, 0.9, 0.95, 0.99, 0.999\}$. Additionally, we validated the maximal number of parents, where we used 1 or 2 parents per node. When the training set was sufficiently large ($> 1000$ samples) we used 20% of the training samples as validation set. Otherwise, we used 5-fold cross validation. In all cases, we used the same validation set/cross-folds for SVM training and for our algorithms.

For solving MILPs we used the ILOG CPLEX optimizer.[3] For each optimization problem we set a time limit of 2 hours, i.e. if after 2 hours an optimization had not finished, we stopped it and used the best solution found so far. When maximizing the SM, for most datasets an optimal solution was found within these two hours, except for "letter", "satimage", "segment", "soybean-large", "vehicle", "adult", and "spambase". For the datasets "letter" and "soybean-large" the resulting MILPs were too large to return a reasonable solution at all: only the trivial unconnected DAG, found by an internal CPLEX heuristic was returned. However, when maximizing the SBM, a reasonable solution was found in any case. Table 1 shows the worst-case sub-optimality bounds for the "problematic" datasets, according to $100 \frac{\bar{z}-z}{\bar{z}}\%$, where $\bar{z}$ is the B&B upper bound of the margin score, and $z$ is the objective of the best feasible solution. For the generative BNs using the MDL score, we used our formulation (19) without margin constraints and used the linear MDL objective (cf. (3)). We used the same set of parent-sets as for max-margin training, and also cross-validated the number of parents per node (1 or 2 parents). For all datasets an optimal solution was found, where the optimization time was typically under 1 second. The longest optimization time of 716 seconds was needed for "spambase" (58 variables).

The classification results for all datasets are shown in table 2, where the estimated accuracy, together with the 95% confidence intervals is shown (Mitchell, 1997). Table 3 summarizes these results, where the plain and boldface numbers are the number of times a classifier outperforms an other classifier with a significance level of 68% and 95%, respectively. For the significance tests, we used a one-sided paired t-test for the datasets with cross-validation, and a one-sided binomial test for the other datasets. We see that the max-margin structures SM and SBM perform clearly better

---
[3]ILOG CPLEX is freely available for non-commercial research under http://www.ibm.com/

Exact Maximum Margin Structure Learning of Bayesian Networks

Table 1. Relative worst-case sub-optimality bounds for SM and SBM in %.

| Method | letter | satimage | segment | soybean-large | vehicle | adult | spambase |
|--------|--------|----------|---------|---------------|---------|-------|----------|
| SM     | ∞      | 7.60     | 16.58   | ∞             | 4.57    | 2.42  | 1.07     |
| SBM    | 1.99   | 15.04    | 0.00    | 0.00          | 0.00    | 0.58  | 1.07     |

Table 2. Mean classification accuracy with 95% confidence intervals for 31 datasets. NB: Naive Bayes, TAN: tree-augmented naive Bayes, MDL: generative Bayesian network trained with MDL criterion, SM: soft margin (this paper), SBM: binary soft margin (this paper), SVM: support vector machine with Gaussian kernel. Datasets "letter" and "soybean-large" had too many samples/class values for SM to return a result.

| Dataset | NB | TAN 1 | MDL | SM | SBM | SVM |
|---------|----|----|-----|----|-----|-----|
| australian | 86.50 ± 2.76 | 86.08 ± 2.43 | 85.34 ± 2.93 | 85.77 ± 3.02 | 85.77 ± 3.02 | **86.80 ± 2.85** |
| breast | 97.39 ± 1.64 | 97.10 ± 1.98 | 96.95 ± 1.51 | 97.24 ± 1.97 | 97.24 ± 1.97 | **97.69 ± 1.95** |
| chess | 87.34 ± 2.00 | 93.71 ± 1.46 | 91.74 ± 1.65 | 95.50 ± 1.24 | 95.50 ± 1.24 | **98.50 ± 0.73** |
| cleve | 83.78 ± 6.60 | 80.74 ± 6.77 | 82.43 ± 6.26 | **84.13 ± 6.01** | **84.13 ± 6.01** | 82.76 ± 6.24 |
| corral | 87.37 ± 9.73 | 90.57 ± 9.09 | **100.00 ± 0.00** | **100.00 ± 0.00** | **100.00 ± 0.00** | **100.00 ± 0.00** |
| crx | 85.45 ± 6.46 | **86.38 ± 3.60** | 86.08 ± 3.92 | 84.68 ± 3.34 | 84.68 ± 3.34 | 86.37 ± 1.65 |
| diabetes | 75.12 ± 4.28 | 74.08 ± 4.11 | 75.38 ± 3.58 | **75.52 ± 3.62** | **75.52 ± 3.62** | 73.84 ± 4.28 |
| flare | 80.20 ± 6.68 | **82.92 ± 4.78** | 82.64 ± 2.60 | 81.60 ± 3.29 | 81.60 ± 3.29 | 82.45 ± 1.27 |
| german | **75.00 ± 3.90** | 73.70 ± 3.30 | 72.60 ± 2.34 | 73.60 ± 3.60 | 73.60 ± 3.60 | 73.40 ± 3.81 |
| glass | 70.83 ± 8.53 | 69.94 ± 7.88 | 70.33 ± 9.27 | 70.70 ± 8.49 | **72.69 ± 9.92** | 71.33 ± 3.91 |
| glass2 | 81.62 ± 5.17 | 80.27 ± 6.87 | 81.62 ± 5.17 | **82.77 ± 5.44** | **82.77 ± 5.44** | 80.27 ± 8.36 |
| heart | **84.07 ± 6.82** | 82.22 ± 5.29 | 83.33 ± 7.63 | 83.33 ± 6.90 | 83.33 ± 6.90 | 83.70 ± 6.97 |
| hepatitis | 86.33 ± 8.33 | 89.00 ± 9.66 | 85.33 ± 6.93 | 89.33 ± 7.40 | 89.33 ± 7.40 | **89.67 ± 11.74** |
| iris | 94.00 ± 1.85 | **95.33 ± 2.27** | 94.00 ± 1.85 | 93.33 ± 2.93 | 93.33 ± 2.93 | 93.33 ± 2.93 |
| letter | 73.50 ± 1.22 | 85.60 ± 0.97 | 74.86 ± 1.20 | N/A | 86.64 ± 0.94 | **97.78 ± 0.41** |
| lymphography | **87.09 ± 11.49** | 87.09 ± 10.89 | 81.90 ± 10.62 | 83.59 ± 13.17 | 85.39 ± 11.36 | 83.10 ± 2.09 |
| mofn-3-7-10 | 86.72 ± 2.08 | 90.82 ± 1.77 | 88.28 ± 1.97 | 92.19 ± 1.64 | 92.19 ± 1.64 | **100.00 ± 0.00** |
| pima | 75.78 ± 1.15 | 75.39 ± 2.58 | **76.04 ± 0.76** | 75.40 ± 1.84 | 75.40 ± 1.84 | 75.66 ± 1.92 |
| satimage | 81.40 ± 1.71 | 87.45 ± 1.45 | 85.00 ± 1.56 | 88.50 ± 1.40 | 87.45 ± 1.45 | **91.35 ± 1.23** |
| segment | 90.65 ± 2.06 | 92.73 ± 1.83 | 90.65 ± 2.06 | 95.97 ± 1.39 | 95.32 ± 1.49 | **97.53 ± 1.10** |
| shuttle-small | 98.81 ± 0.48 | 99.48 ± 0.32 | 99.53 ± 0.30 | **99.79 ± 0.20** | 99.64 ± 0.27 | 99.43 ± 0.34 |
| soybean-large | 91.62 ± 1.20 | 92.03 ± 2.11 | 88.12 ± 2.84 | N/A | **93.98 ± 1.74** | 93.76 ± 2.14 |
| vehicle | 59.13 ± 2.99 | 69.14 ± 2.92 | **69.88 ± 3.43** | 69.26 ± 0.46 | 69.03 ± 1.85 | 68.74 ± 4.20 |
| vote | 90.17 ± 4.41 | 94.07 ± 3.00 | 93.85 ± 3.71 | **95.65 ± 3.08** | **95.65 ± 3.08** | 94.48 ± 3.29 |
| waveform-21 | 78.85 ± 1.17 | 79.21 ± 1.16 | 78.85 ± 1.17 | 79.43 ± 1.16 | 79.43 ± 1.16 | **79.85 ± 1.15** |
| abalone | 56.53 ± 3.02 | 62.34 ± 2.95 | 62.15 ± 2.96 | **64.18 ± 2.92** | 63.31 ± 2.94 | 62.54 ± 2.95 |
| adult | 83.75 ± 0.59 | 85.11 ± 0.57 | 85.29 ± 0.57 | **86.39 ± 0.55** | **86.39 ± 0.55** | 84.84 ± 0.57 |
| car | 82.96 ± 3.07 | 86.43 ± 2.80 | 83.48 ± 3.04 | 92.52 ± 2.15 | 91.65 ± 2.26 | **97.57 ± 1.26** |
| mushroom | 95.16 ± 0.81 | 99.96 ± 0.07 | 99.85 ± 0.14 | **100.00 ± 0.00** | **100.00 ± 0.00** | **100.00 ± 0.00** |
| nursery | 90.05 ± 0.73 | 92.67 ± 0.63 | 89.72 ± 0.74 | 94.58 ± 0.55 | 94.71 ± 0.55 | **99.77 ± 0.12** |
| spambase | 89.65 ± 1.24 | 92.65 ± 1.07 | 92.70 ± 1.06 | **94.48 ± 0.93** | **94.48 ± 0.93** | 93.96 ± 0.97 |

than NB, TAN and MDL, since a max-margin structure outperforms the other BNs at least 17 times (11 times significantly), while NB, TAN and MDL outperform a max-margin structure maximal 7 times (2 times significantly). On the other hand, the SVM outperforms the max-margin structures at least 11 times (6 times significantly), while a max-margin structure outperforms the SVM maximal 9 times (3 times significantly). Therefore, there is a trend in favor of the SVM, although the results for the max-margin structures are in the same range. As already mentioned, the BNs with max-margin structure still use generative parameters. Therefore, the resulting models still consistently approximate the empirical data distribution and are amenable for other inference tasks than classification. Additionally, we could train the parameters of the resulting BNs in a discriminative way, to further improve classification results (Guo et al., 2005; Pernkopf et al., 2012). We plan to address this in future work.

Table 3. Number of times where classifier in row outperformed classifier in column. Plain letters: significance level 68%. Boldface letters: significance level 95%.

| \ | NB | TAN | MDL | SM | SBM | SVM |
|---|----|-----|-----|----|-----|-----|
| NB | — | 7/**2** | 9/**3** | 7/**2** | 7/**1** | 7/**0** |
| TAN | 20/**16** | — | 16/**10** | 4/**0** | 4/**0** | 4/**0** |
| MDL | 15/**12** | 8/**3** | — | 5/**1** | 5/**1** | 7/**1** |
| SM | 18/**14** | 17/**13** | 19/**11** | — | 4/**2** | 8/**3** |
| SBM | 21/**16** | 18/**14** | 21/**12** | 3/**0** | — | 9/**2** |
| SVM | 20/**17** | 15/**10** | 16/**10** | 11/**6** | 12/**7** | — |

## 7. Conclusion

We proposed an exact method for the combinatorial problem of finding a BN structure maximizing the probabilistic soft margin, extending previous methods for exact generative BN structure learning. We demonstrated the applicability of our methods on small and medium sized datasets and produced promising results. Having an exact algorithm is valuable – although the problem is NP-hard. Firstly, it is important to address those datasets were the problem turns out to be tractable. Secondly, a key feature of the methods presented in this paper is that they provide any-time solutions, i.e. when the problem turns out to be infeasible, they still return an approximation together with a worst-case estimate of sub-optimality. Therefore, these methods can also provide satisfying results on more difficult problems. Furthermore, exact methods provide interesting theoretical insights into the problem nature, and possibly motivate new heuristics and approximations for inexact structure learning.

## Acknowledgments

This work was supported by the Austrian Science Fund (project number P22488-N23).